\documentclass[letterpaper, 10 pt, conference]{ieeeconf}  
\IEEEoverridecommandlockouts 
\newif\ifblindreview

\usepackage{graphics} 
\usepackage{epsfig} 
\usepackage{mathptmx} 
\usepackage{times} 
\usepackage{amsmath} 
\usepackage{amssymb}  
\usepackage{booktabs}    
\usepackage{multirow}    
\usepackage{array}      
\usepackage{makecell}
\title{\LARGE \bf
Design and Fabrication of Origami-Inspired\\ Knitted Fabrics for Soft Robotics
}

\ifblindreview
  \author{\IEEEauthorblockN{Authornames and affiliations withheld for double anonymous review}
  \IEEEauthorblockA{}
  }
\else
  \author{
    Sehui Jeong$^{1}$, 
    Magaly C. Aviles$^{1}$,
    Athena X. Naylor$^{1}$ \\
    Cynthia Sung$^{2}$,
    Allison M. Okamura$^{1}$%
    \thanks{$^{1}$Department of Mechanical Engineering, Stanford University, Stanford, CA 94305, USA.}%
    \thanks{$^{2}$Department of Mechanical Engineering, University of Pennsylvania, Philadelphia, PA 19104, USA.}
}
\fi

\usepackage{xcolor}

\begin{document}

\maketitle
\thispagestyle{empty}
\pagestyle{empty}

\begin{abstract}
Soft robots employing compliant materials and deformable structures offer great potential for wearable devices that are comfortable and safe for human interaction. However, achieving both structural integrity and compliance for comfort remains a significant challenge. In this study, we present a novel fabrication and design method that combines the advantages of origami structures with the material programmability and wearability of knitted fabrics. We introduce a general design method that translates origami patterns into knit designs by programming both stitch and material patterns. The method creates folds in preferred directions while suppressing unintended buckling and bending by selectively incorporating heat fusible yarn to create rigid panels around compliant creases. We experimentally quantify folding moments and show that stitch patterning enhances folding directionality while the heat fusible yarn (1) keeps geometry consistent by reducing edge curl and (2) prevents out-of-plane deformations by stiffening panels. We demonstrate the framework through the successful reproduction of complex origami tessellations, including Miura-ori, Yoshimura, and Kresling patterns, and present a wearable knitted Kaleidocycle robot capable of locomotion. The combination of structural reconfigurability, material programmability, and potential for manufacturing scalability highlights knitted origami as a promising platform for next-generation wearable robotics.

\end{abstract}

\section{INTRODUCTION}
Soft robots operate effectively in human environments by conforming to their surroundings using their material compliance~\cite{rus2015design,kim2013soft,yasa2023overview}. Morphing geometries further expand the range of achievable deformation, which is difficult to obtain through material stretch alone \cite{meloni2021engineering,misseroni2024origami}. In particular, origami structures, derived from the art of paper folding, transform sheet materials into three-dimensional structures that allow complex and large shape change using scalable planar fabrication processes. Leveraging this morphology change, origami-inspired robots can provide complex and controlled motions such as grasping, crawling, and propulsion \cite{rus2018spotlight, feshbach2023curvequad, yang2021origami}.

Origami achieves folding through rigid panels connected by flexible hinges. Conventional fabrication of origami robots relies on laser cutting, sequential layer stacking, or manual folding \cite{felton2014method, hawkes2010programmable}. However, foldable hinges, created by laser cutting and modifying local thickness, often fail due to material degradation because the thin parts have weak durability. Multimaterial composites made by layer stacking often fail at interfaces because of a stiffness mismatch due to delamination or poor adhesion. Additive manufacturing has enabled higher geometric fidelity and programmable structures, yet, achieving robust interfaces between soft and rigid parts remains as major challenge~\cite{xue2025origami}.

\begin{figure}
\centering
    \includegraphics[width=\linewidth]{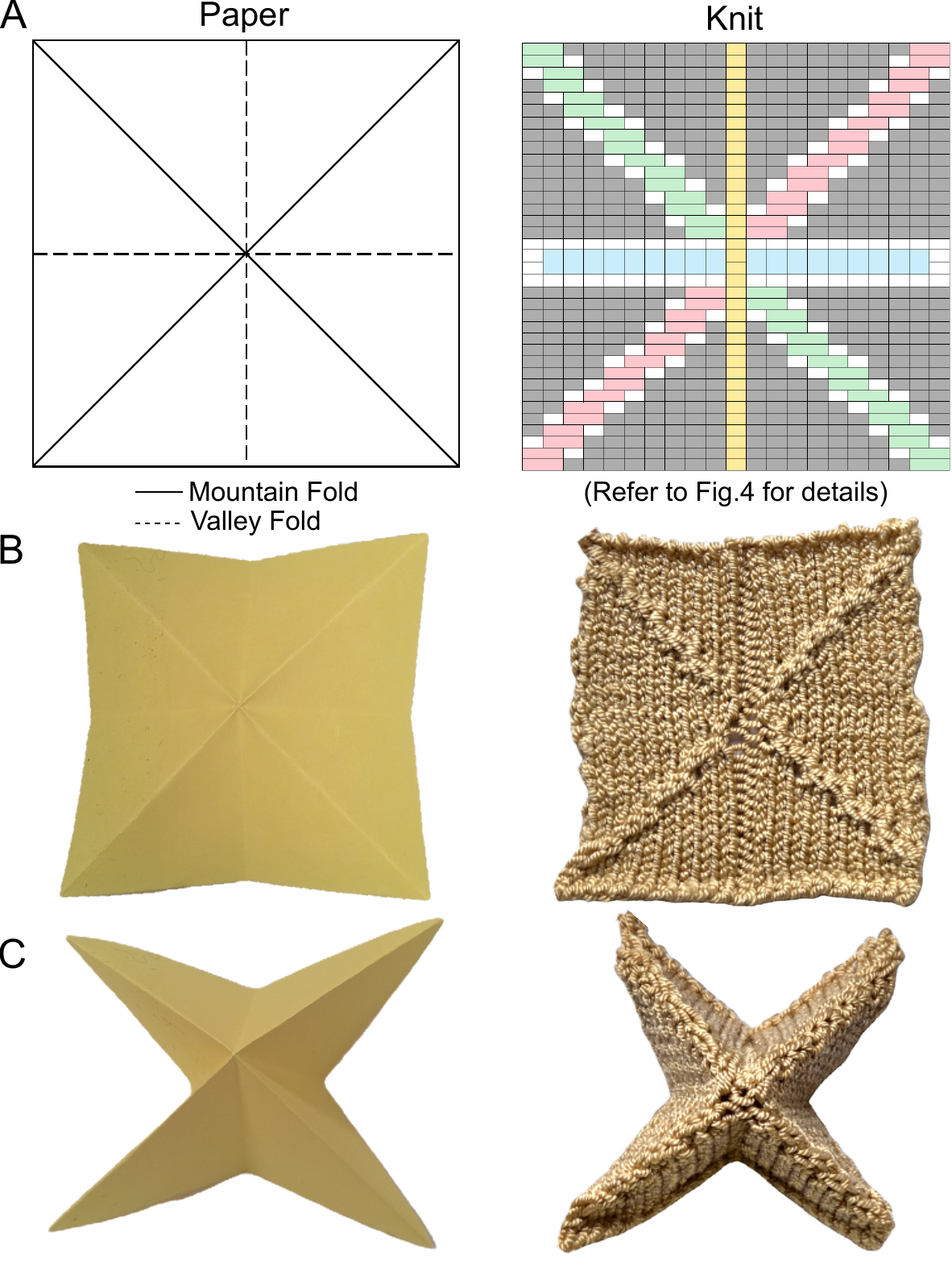}
    \caption{Translation of a paper origami pattern into a knitted origami structure.
(A) Paper origami preliminary base pattern (left), and translated knit origami pattern (right). In the knit pattern, different colors refer to different stitch types, where the details and the enlarged pattern are provided in Figure~\ref{fig:Fig4}.
(B) Deployed configurations of paper and knitted origami.
(C) Collapsed configurations of paper and knitted origami.}
    \label{fig:Fig1}
\end{figure}

Knitted fabrics have been adopted in soft robotics for actuators, mobile robots, and wearable devices \cite{narayanan2019visual, granberry2017active, kim2022knitskin}. Industrial knitting machines provide spatial control of stitch patterns and materials, expanding the design space \cite{hofmann2020knitgist, luo2022digital}. As an example of the design capability, Figure~\ref{fig:Fig1} provides a knit design involving different types of stitches and yarns. The yarn material governs the mechanical properties, while stitch type further modulates stiffness and anisotropy by changing the topology and symmetry of yarn entanglement~\cite{du2024haptiknit, tan2023geometric,du2025fiber, singal2024programming}. Thus, the combination of stitch pattern and yarn properties in programmed machine knitting enables tuning of stiffness and nonlinear response and supports designs that benefit from distributed stiffness. 

Previous works developed self-folding auxetic knitted textiles by alternating knit and purl stitches, extending pleats in fashion to structural design of knits~\cite{hu2011development, knittel2015self, liu2010negative, luan2020auxetic}. These works exploit mirror symmetry between knit and purl to generate folding moments along the interface. Because this mechanism relies on simple alternation, this approach to inducing folds is strictly limited to origami patterns where mountain and valley folds alternate. Therefore, a central challenge remains in the lack of a general design framework for knitted fabrics to achieve structural reconfiguration using arbitrary origami patterns.

In this study, we overcome this challenge by presenting a novel fabrication and design method that combines the advantages of origami structure and knitted fabrics. First, we introduce a general design method that translates origami patterns into knit designs by specifying stitch and material patterns. The method creates folds in a preferred direction while suppressing buckling and unintended bending. Second, we experimentally quantify folding moments to analyze the folding behavior. As expected, the proposed method shows both reduced bending moments in the prescribed direction and improved folding directionality. This design also results in more accurate geometry by preventing edge curling and out-of-plane deformation. Finally, we demonstrate knitted origami tessellations and a wearable Kaleidocycle robot. The robot benefits from an origami geometry that enables continuous rotation and radius modulation for locomotion along slender bodies, and the knitted fabric preserves wearability. The demonstrations show the potential of knit origami for wearable devices by integrating structural reconfiguration with user comfort provided by knit.

\section{Methods}
Knitted fabrics can have spatially controlled stiffness and yarn topology by varying yarn materials and stitch types. In this section, we introduce a design method to convert origami pattern in to knit design. Figure~\ref{fig:Fig1} illustrates the translation of a preliminary base pattern from a paper origami to knit. The pattern consists of diagonal mountain folds and horizontal and vertical valley folds along the center lines. The knit design uses knit, purl, tuck, and twist stitches, and selectively incorporates heat fusible yarn in regions intended to remain rigid. Comparison of the deployed and folded configurations of paper and knit origami shows that the paper origami pattern is successfully transferred to knit. The following sections explain the design principles and processes and the resulting patterns in detail.

All samples in this study were fabricated on a Silver Reed SK840 knitting machine with SRP60N ribber and AG24 intarsia carriage. The ribber adds a second needle bed, enabling the creation of purl stitches. The intarsia carriage facilitates complex multimaterial patterning by allowing distinct yarns within the same row.

\begin{figure}
    \centering
    \includegraphics[width=\linewidth]{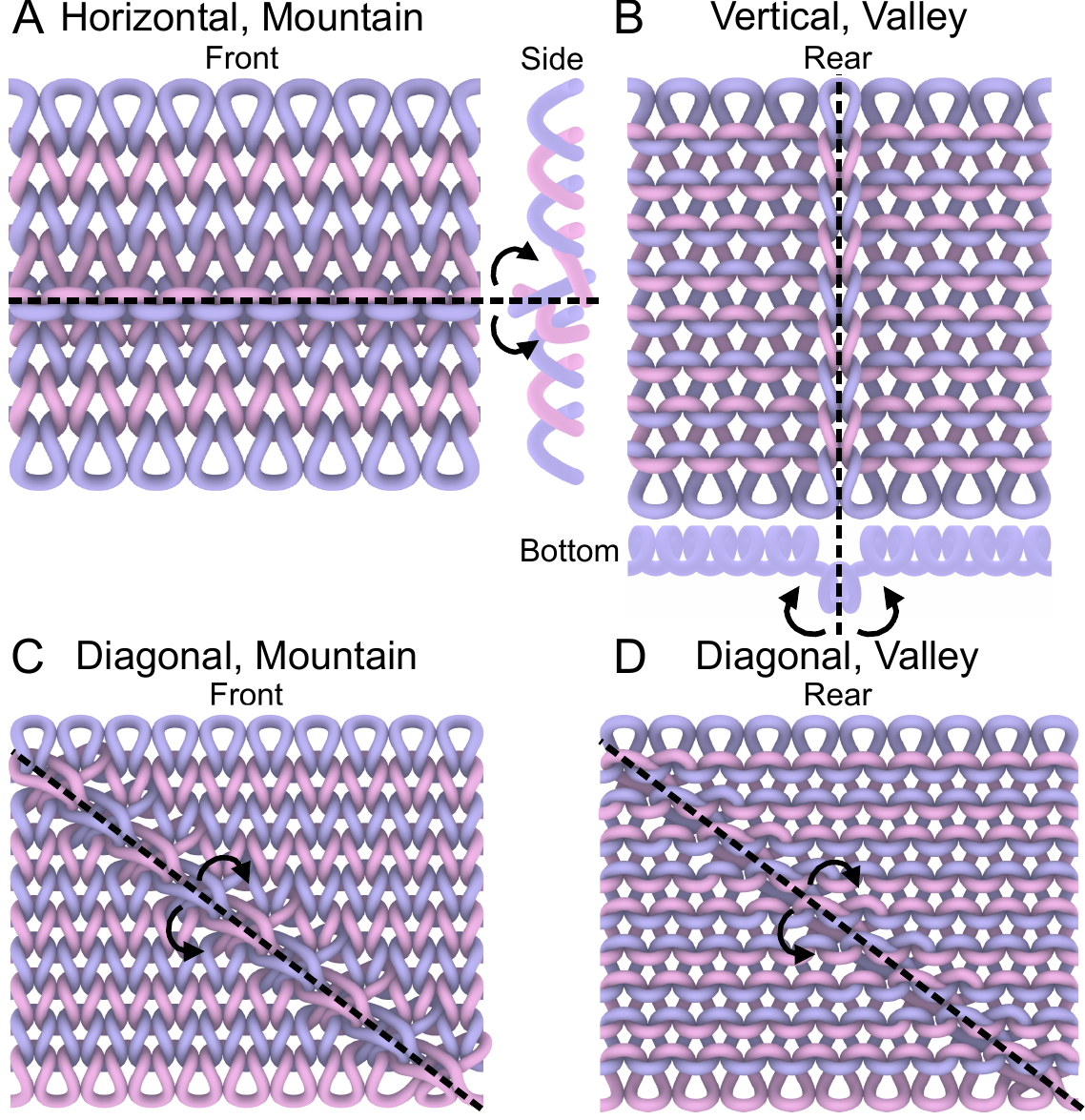}
    \caption{Graphical representation of stitch patterns that create folds (dashed lines) and the resulting folding moments due to asymmetry and local thickness changes created by stitch patterns (arrows).
(A) Front view of a horizontal mountain fold. (B) Rear view of a vertical valley fold. (C) Front view of a backward-slant diagonal ($\diagdown$) mountain fold. (D) Rear view of a forward-slant diagonal ($\diagup$) valley fold.}
    \label{fig:Fig2}
\end{figure}
\subsection{Stitch Patterns}

This section investigates stitch types that create folds in prescribed directions. Figure~\ref{fig:Fig2} presents a rendered graphic of folds generated by these stitch patterns, with arrows indicating the locations and mechanisms of the structurally induced folding moments.
 
 Knit and purl are the two fundamental stitches. A knit stitch is created by pulling a new loop through an existing loop from the front, producing a V-shaped stitch on the front face. A purl stitch is formed by drawing the yarn through from the back, producing a horizontal ridge on the front face. Knit and purl are mirror images of each other. In this study, we use a jersey pattern as a baseline, which is composed entirely of knit stitches.

Garter and rib patterns, where knits and purls alternate along the course and wale directions, exhibit increased anisotropy, with the fabric softer in those directions due to bending moments induced by structural asymmetry~\cite{singal2024programming}. This motivates selective use of purl stitches to induce folding. We find that a row of purl stitches aligned with the course direction forms a mountain fold, whereas a column of purl stitches aligned with the wale direction forms a valley fold.

\begin{figure*}[t]
    \centering
    \includegraphics[width=1\textwidth]{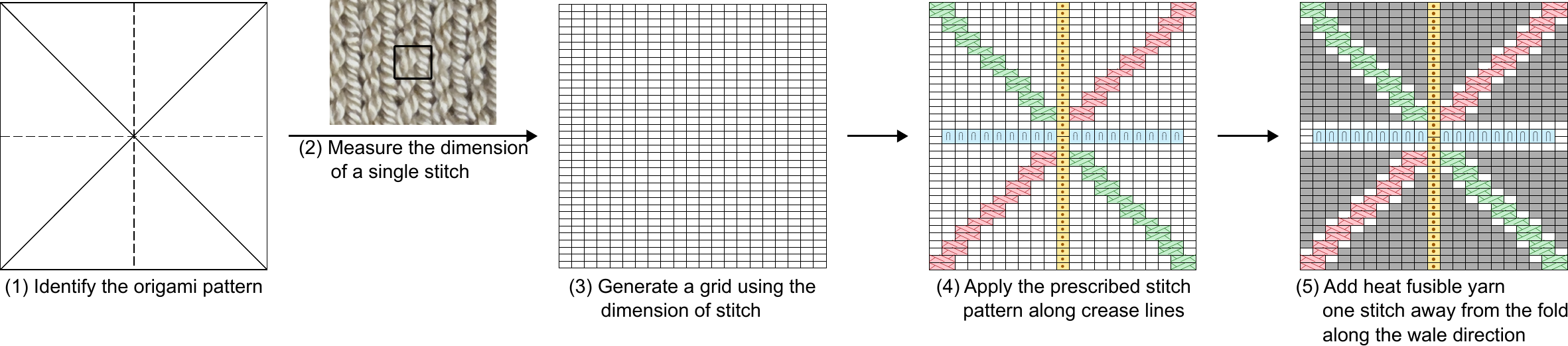}
    \caption{A schematic of the framework to translate an origami pattern into a knit structure. The process consists of five steps: (1) identify the target origami pattern, (2) measure the dimensions of a single knit stitch, (3) generate a stitch-level grid, (4) apply the prescribed stitch types (e.g., knit, purl, tuck, twist) along the crease lines, and (5) selectively add heat-fusible yarn to program the fold behavior. The details of the pattern in step 5 are shown in Figure~\ref{fig:Fig4}}
    \label{fig:Fig3}
\end{figure*}

The asymmetry of the knit stitch can be amplified with tuck stitches. A tuck stitch is formed by holding a loop on the needle while knitting the next stitch. This operation effectively adds yarn and increases local thickness. It shifts the neutral axis for out-of-plane bending to induce folding. Accordingly, a row of tuck stitches along the course direction produces a valley fold, whereas a column of tuck stitches along the wale direction produces a mountain fold.

Diagonal folds can be created with twist stitches, in which two adjacent stitches cross to form a twist. Twisting stretches one side of the stitch, creating local bumps and an asymmetry that generates a moment about the intended fold. Consequently, a twist stitch can act as either a mountain or a valley fold depending on the twisting direction and the direction of propagation, enabling both forward and backward-slant diagonal creases.

\subsection{Materials}
Origami exploits the difference in stiffness between the crease and the surrounding panels \cite{qiu2013kinematic}. The crease line must be sufficiently compliant to undergo bending so that it can be folded and deployed easily, whereas the adjacent panels require sufficient rigidity to remain planar \cite{wheeler2016soft}. Without this stiffness contrast, the entire structure is prone to out-of-plane deformations such as bending and buckling.

A knitting machine can precisely control stiffness by varying materials stitch by stitch. Incorporating heat fusible yarn can locally increase stiffness through post-heating while preserving fabric texture and geometry. According to the previous study, when combined with nylon and Lycra yarn, incorporating heat fusible yarn (Griltech 390 Denier Grilon K85) up to 54.2\% by weight increased stiffness by as much as 4.9 times compared to the base material~\cite{du2024haptiknit}. In this study, we used microfiber acrylic yarn (Alize Diva Stretch Bikini yarn), which consists of 92\% acrylic microfiber and 8\% elastic yarn, to achieve smooth texture and stretchability, with a heat fusible yarn, same as previous study, at 52\% by weight relative to the acrylic yarn.

To achieve the stiffness contrast, heat fusible yarn is selectively incorporated. Rigid panels use acrylic combined with heat fusible yarn, whereas crease lines use acrylic alone. This keeps the creases bendable while stiffening the surrounding panels to suppress twisting, buckling, and unintended out-of-plane bending for structure reliability.

\subsection{Design Method for Translating Origami to Knits}

\begin{figure}
\centering
    \includegraphics[width=\linewidth]{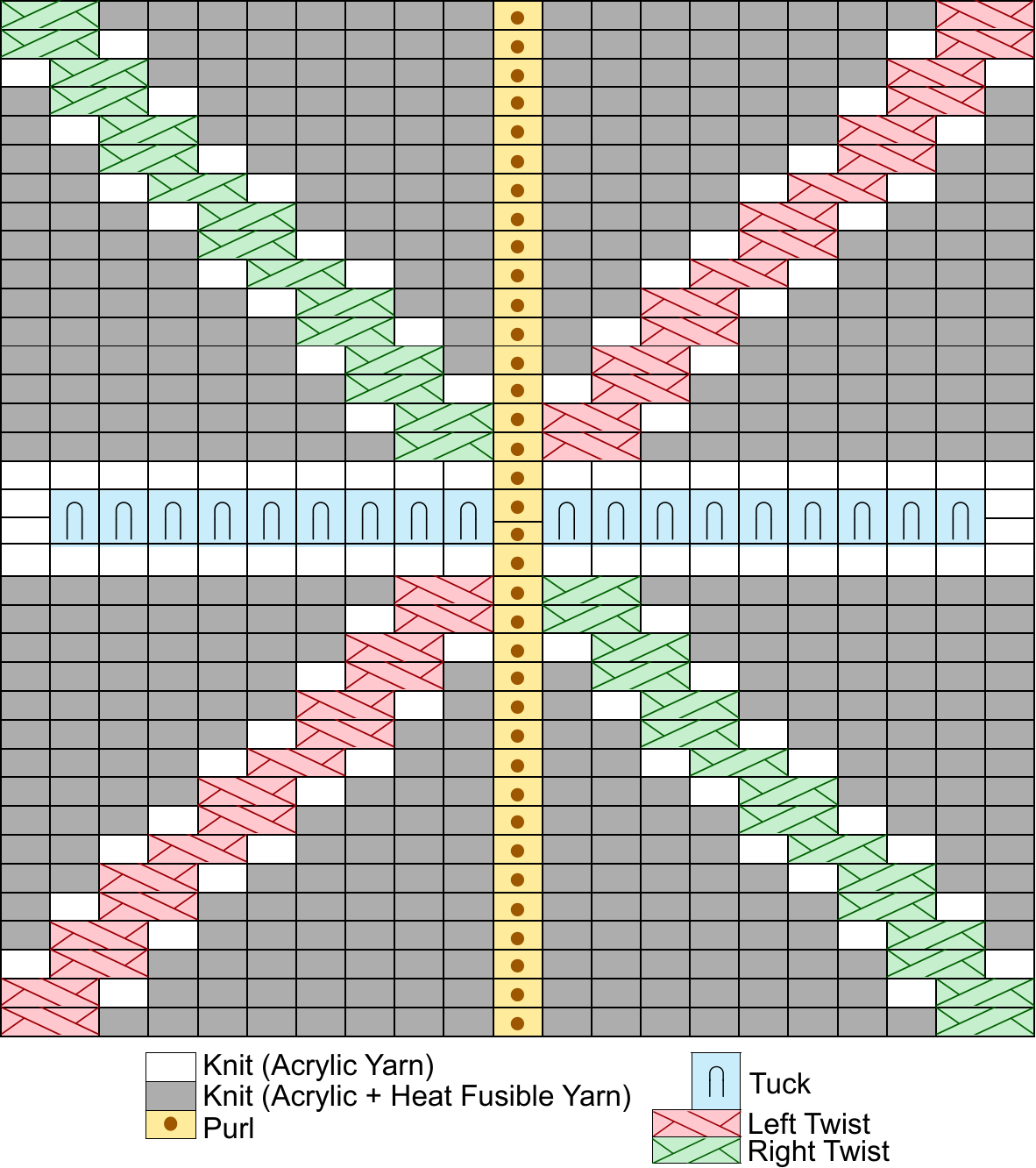}
    \caption{Detailed patterns corresponding to Figure~\ref{fig:Fig1}(A). Symbols follow the \textit{Knitting Font Collection}. The colored cells represent distinct stitch types: gray cells correspond to knit stitches with acrylic and heat-fusible yarn. All other colors represent stitches made with acrylic yarn only: white (knit), yellow (purl), blue (tuck), red (left twist), and green (right twist).}
    \label{fig:Fig4}
\end{figure}

Figure~\ref{fig:Fig3} shows the design principle for translating paper origami to knit, for the pattern shown in Figure~\ref{fig:Fig1} with the detailed pattern presented in Figure~\ref{fig:Fig4}.

First, we identify the origami pattern, the mountain and valley folds, and their orientations. We then measure the stitch dimensions on plain knitted fabric produced with the targeted material and manufacturing parameters. These dimensions are measured after removal from the knitting needle in the relaxed state. We then generate a grid whose cell size matches the stitch dimensions to map the origami pattern onto the knit domain with minimal distortion. 

We then apply the prescribed stitch type along the crease lines to encode fold direction and orientation. Here, a tuck is two cells tall because it is created by holding one stitch and pulling two stitches in the next row. A twist is two cells wide because it is made by swapping two adjacent loops. 

When diagonal folds intersect, we leave the intersection as knit because twisting subsequent loops overly stretches the yarn and often causes yarn failure. When horizontal and vertical folds intersect, if the required stitches are identical for the two directions, we use that stitch. If they differ, we choose purl to preserve structural continuity across the intersection. The structural contrast between knit and purl is greater than the difference between knit and tuck, which is mostly a local thickness change.

Finally, we add heat fusible yarn to stitches at least one stitch away from the fold along the wale direction. A stitch is formed by pulling new yarn into an existing loop, so stitches that are adjacent in the vertical direction influence one another. This placement keeps the fold soft while adding stiffness to the panels. Figure~\ref{fig:Fig4} shows the detailed knit design, which was translated from the preliminary base pattern in Figures~\ref{fig:Fig1} and \ref{fig:Fig3}.

\section{Folding Characterization}

This section experimentally characterizes the folding behavior of knitted origami fabricated using the method of Section $\mathrm{II}$ by measuring the folding moment and evaluating the effects of stitch patterns and heat fusible yarn. Here, patterns refer to crease lines translated into stitch designs in acrylic yarn according to our framework. Heat fusible yarn is applied in regions at least one stitch away from the fold to add stiffness while keeping the crease soft.

To analyze the folding behavior, we evaluate four distinct fabric configurations. We compare a baseline jersey fabric (acrylic yarn only, no pattern) against a fabric with only stitch patterns applied. We then compare these to jersey fabric with heat fusible yarn on the panels and the other combining both stitch patterns and heat fusible yarn, representing our complete design framework.

Each configuration is tested across four orientations (horizontal, vertical, forward-slant diagonal ($\diagup$), and backward-slant diagonal ($\diagdown$)) and two fold types (mountain and valley). The baseline jersey fabric does not contain any programmed fold, so we report a single representative value for this control case. For the unpatterned configurations (baseline jersey and jersey with heat fusible yarn), mountain forward is identical to valley backward, and vice versa. For the unpatterned, heat fusible fabric, the two diagonal orientations are also identical. After removing redundancies due to symmetry, we consider 20 total experimental conditions.

\begin{figure}[t!]
    \centering
    \includegraphics[width=\linewidth]{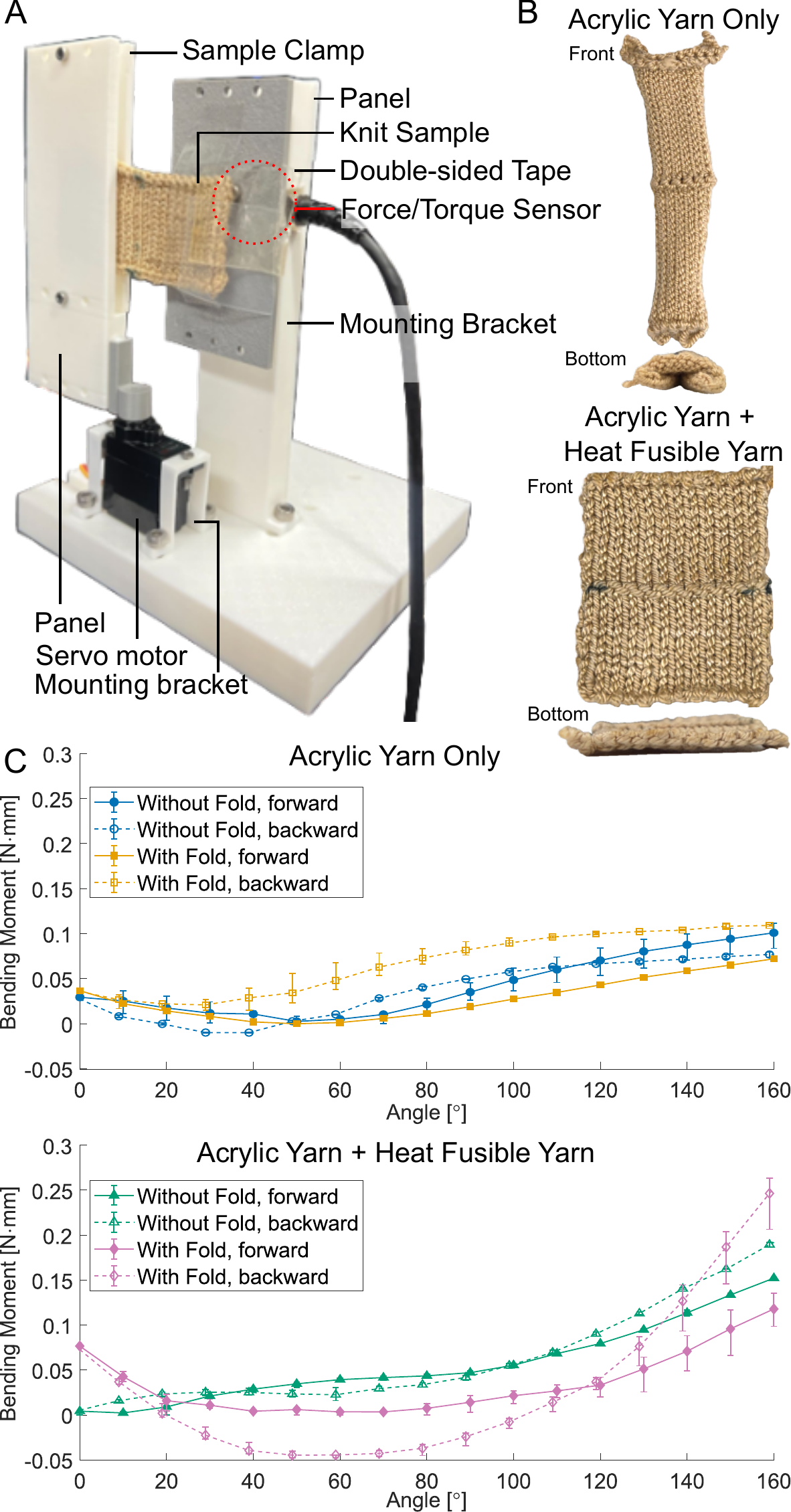}
    \caption{(A) Photograph of the experimental setup. (B) Front and bottom views of sample with a horizontal mountain fold without heat fusible yarn (top) and with heat fusible yarn (bottom). Without heat fusible yarn, the edge of sample curl due to inherent asymmetry of the yarn geometry. (C) Experimental results for the horizontal mountain fold in forward and backward directions, normalized by crease length. The backward results are slightly shifted to avoid overlap with the forward results. The error bars indicate the minimum and maximum of measured values. (Top) Comparison between the jersey fabric and the patterned fabric made of acrylic yarn only. (Bottom) Comparison between the jersey fabric and the patterned fabric with  heat fusible yarn.}
    \label{fig:Fig5}
\end{figure}

The experimental setup in Figure~\ref{fig:Fig5}(A) characterizes the folding moment of knitted origami at prescribed fold angles. The apparatus was adapted from \cite{xiao2023folding} and consists of a force sensor mount, motor supports, a sensor-mounted panel, and an actuated panel. During folding, the reaction force is measured with an ATI Nano17 force/torque sensor. The folding moment per unit crease length \(M\) is computed as \(M=Fd/L\), where \(F\) is the measured normal force, \(d\) is the perpendicular distance from the motor rotation axis to the center of the sensor, and \(L\) is the length of the crease line aligned with the rotation axis. Moments are reported per unit crease length to allow comparison across specimens of different sizes.

The sample is mounted between two panels on the actuated side. On the sensor side, a layer of double sided tape is applied to align the sample so that the crease line matches the motor rotation axis. The edges are secured with clear tape to prevent edge curling. The servo motor is commanded to rotate from \(0^\circ\) to \(180^\circ\) in \(10^\circ\) increments while the sensor records the normal force. The forward direction is defined as the programmed direction of the fold. For jersey fabric, which serves as the baseline without the proposed patterns, the front side is where V-shaped loops are visible. The test is repeated three times for each condition.

For concise comparison, we define a directionality ratio
\begin{equation}
R \;=\; \frac{M_{\text{forward}}}{M_{\text{backward}}}\,,
\end{equation}
where $M_{\text{forward}}$ and $M_{\text{backward}}$ are the peak moments measured in the forward and backward directions. Note that larger directionality ratio indicates preferred folding in programmed direction. As an example, results for a horizontal mountain fold are shown in Figure~\ref{fig:Fig5}(C), along with a visual qualitative comparison. The error bars indicate the minimum and maximum values. The full results appear in Table~\ref{tab:fold-results}.

\begin{table}[t!]
  \centering
  \caption{Folding moments $M_{\text{forward}}$ and $M_{\text{backward}}$ and the directionality ratio $R$ for knitted origami specimens}

  \scriptsize
  \begin{tabular}{c|c|ccc|ccc}
    \toprule
    Fold  & Crease & \multicolumn{3}{c|}{Jersey Fabric} & \multicolumn{3}{c}{Patterned Fabric} \\
    \cmidrule(lr){3-5} \cmidrule(lr){6-8}
        Type      & Orientation   & \makecell{$M_{\text{forward}}$ \\ {}[\tiny$\mathrm{N\cdot mm / mm}$]}  & \makecell{$M_{\text{backward}}$ \\ {}[\tiny$\mathrm{N\cdot mm / mm}$]} & $R$ & \makecell{$M_{\text{forward}}$ \\ {}[\tiny$\mathrm{N\cdot mm / mm}$]} & \makecell{$M_{\text{backward}}$ \\ {}[\tiny$\mathrm{N\cdot mm / mm}$]} & $R$ \\
    \midrule
    \multicolumn{8}{c}{\textbf{Material: Acrylic Yarn on Folds and Panels}} \\
    \midrule
    \multirow{4}{*}{\makecell{Mountain \\ {}Fold}}
      & Horizontal           & 0.046 & 0.053 & 1.16 & 0.032 & 0.041 & 1.29 \\
      & Vertical             & 0.046 & 0.053 & 1.16 & 0.045 & 0.17 & 3.86 \\
      & Diagonal ($\diagup$)         & 0.046 & 0.053 & 1.16 & 0.025 & 0.051 & 2.04 \\
      & Diagonal ($\diagdown$) & 0.046 & 0.053 & 1.16 & 0.027 & 0.04 & 1.50 \\
      \cmidrule(lr){1-8}
    \multirow{4}{*}{\makecell{Valley \\ {}Fold}}
      & Horizontal           & 0.053 & 0.046 & 0.86 & 0.057 & 0.16 & 2.83 \\
      & Vertical             & 0.053 & 0.046 & 0.86 & 0.065 & 0.092 & 1.41 \\
      & Diagonal ($\diagup$)         & 0.053 & 0.046 & 0.86 & 0.028 & 0.029 & 1.02 \\
      & Diagonal ($\diagdown$) & 0.053 & 0.046 & 0.86 & 0.034 & 0.052 & 1.54 \\
    \midrule
    \multicolumn{8}{c}{\textbf{Material: Acrylic on Folds and Acrylic + Heat Fusible Yarn on Panels}} \\
    \midrule
    \multirow{4}{*}{\makecell{Mountain \\ {}Fold}}
      & Horizontal           & 0.081 & 0.070 & 0.87 & 0.060 & 0.16 & 2.69 \\
      & Vertical             & 0.029 & 0.120 & 4.38 & 0.067 & 0.21 & 3.18 \\
      & Diagonal ($\diagup$)         & 0.065 & 0.057 & 0.87 & 0.041 & 0.047 & 1.13 \\
      & Diagonal ($\diagdown$) & 0.065 & 0.057 & 0.87 & 0.031 & 0.066 & 2.17 \\
      \cmidrule(lr){1-8}
    \multirow{4}{*}{\makecell{Valley \\ {}Fold}}
      & Horizontal           & 0.070 & 0.081 & 1.14 & 0.028 & 0.056 & 1.99 \\
      & Vertical             & 0.120 & 0.029 & 0.23 & 0.068 & 0.071 & 1.05 \\
      & Diagonal ($\diagup$)         & 0.057 & 0.065 & 1.14 & 0.028 & 0.13 & 4.56 \\
      & Diagonal ($\diagdown$) & 0.057 & 0.065 & 1.14 & 0.050 & 0.085 & 1.71 \\
    \bottomrule
  \end{tabular}
  \label{tab:fold-results}
\end{table}

Comparing jersey and patterned fabrics using acrylic yarn only, patterned fabrics require lower moments to fold forward in most types. This reduction is consistent with the intended imbalance introduced by purl, tuck, and twist lines that create structural asymmetry or local thickness contrast along the crease. Exceptions occur in horizontal and vertical valley folds, where moments are slightly larger with the pattern than without. This is because the tuck stitch thickens the crease and increases local resistance. For patterned fabrics, backward folds show higher moments than forward folds for the same crease, leading to $R>1$. This reduces the likelihood of snap-through under a force normal to the crease ensuring programmed fold. These results show that the proposed stitch pattern both reduces the folding moment and enhances directionality.

Incorporating heat fusible yarn offers several advantages. It reduces edge curl, which can otherwise hamper folding, and it helps keep the panels planar, preventing out-of-plane buckling. Since the heat fusible yarn is selectively applied to the panels, the crease itself remains soft. However, because knits are formed from a continuous yarn, the stiffened panels are still coupled to the crease. This coupling restricts the stretch of the yarn in the fold that is required to accommodate a complete fold, resulting in an increased folding moment at high angles.

When stitch patterning is used together with heat fusible yarn, $R$ increases as the forward folding moment decreases and the backward moment increases, implying that the programmed folding direction is favored. One exception is vertical mountain folds, where the patterned fabric requires a higher moment than jersey. This is due to increased knitting tension during stitching tuck, which accumulates two loops on a needle and stiffens the crease. It can be mitigated by locally reducing the machine tension with precise control.

Overall, combining stitch patterning with selective use of heat fusible yarn enhances the geometry by removing edge curl, keeps the panels planar and rigid, while improving the folding response by providing compliant hinges that inherently fold in one direction.

\begin{figure}[b!]
    \centering
    \includegraphics[width=\linewidth]{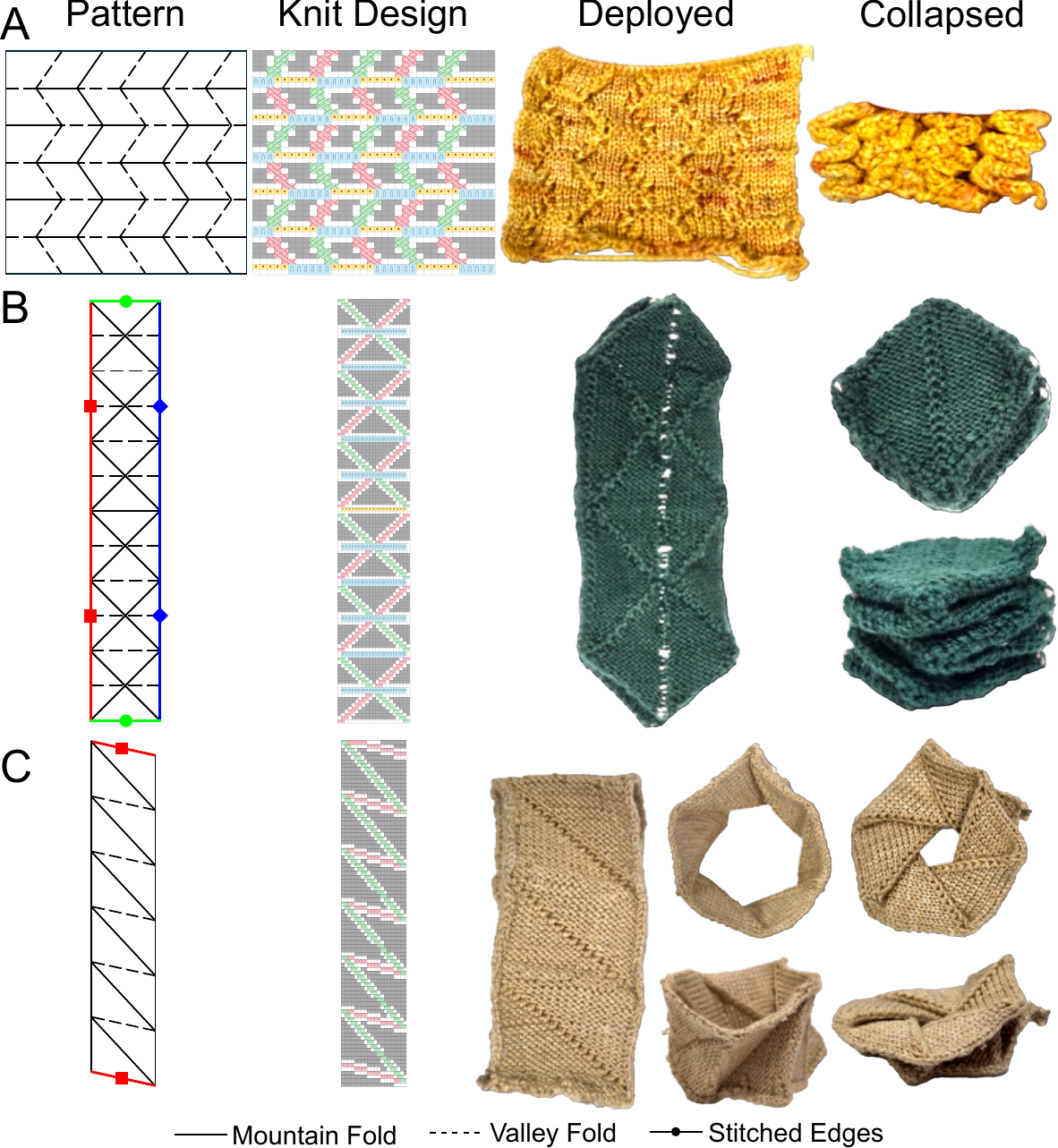}
\caption{Origami tessellations: (A) Miura-ori, (B) Yoshimura, and (C) Kresling.
(First column) Origami crease patterns. Solid and dashed lines indicate mountain and valley folds, respectively. Colored line pairs indicate edges that are stitched together.
(Second column) Corresponding knit designs. The symbols used are defined in Figure~\ref{fig:Fig4}.
(Third column) Photographs of the deployed configurations.
(Fourth column) Photographs of the collapsed configurations.}
    \label{fig:Fig6}
\end{figure}

\section{Applications}
We demonstrate several applications of knit origami by fabricating a range of origami patterns and integrating them into soft robotic systems. These examples validate our strategies for programming knitted textiles. This capability allows for the creation of complex, programmable deployable structures valuable for engineering applications.

\subsection{Origami Tessellations}
Origami tessellation is a pattern formed by repeating folded units that enables large deployable motion. In Figure~\ref{fig:Fig6}, we present knitted versions of the Miura-ori, Yoshimura, and Kresling patterns. For each pattern, we show the original crease pattern together with photographs of the fabricated samples in their deployed and collapsed states.

The Miura-ori pattern is formed by repeating parallelogram units, where diagonal and horizontal creases alternate between mountain and valley folds. This pattern folds with a single degree of freedom and exhibits auxetic behavior, expanding in one direction when stretched in the other. The Yoshimura pattern is constructed from alternating diamond-shaped cells. It exhibits periodic out-of-plane deformations that accommodate both axial shortening and lateral expansion. The Kresling pattern is generated by tessellating triangular facets into a cylindrical configuration, and its deployment couples axial compression with twisting rotation. 

Although these patterns contain different types of folds, our method successfully reproduced each pattern in knitted fabric and achieved the expected collapsing behaviors.

\subsection{Wearable Soft Robot: Knitted Kaleidocycle}
\begin{figure}
    \centering
    \includegraphics[width=\linewidth]{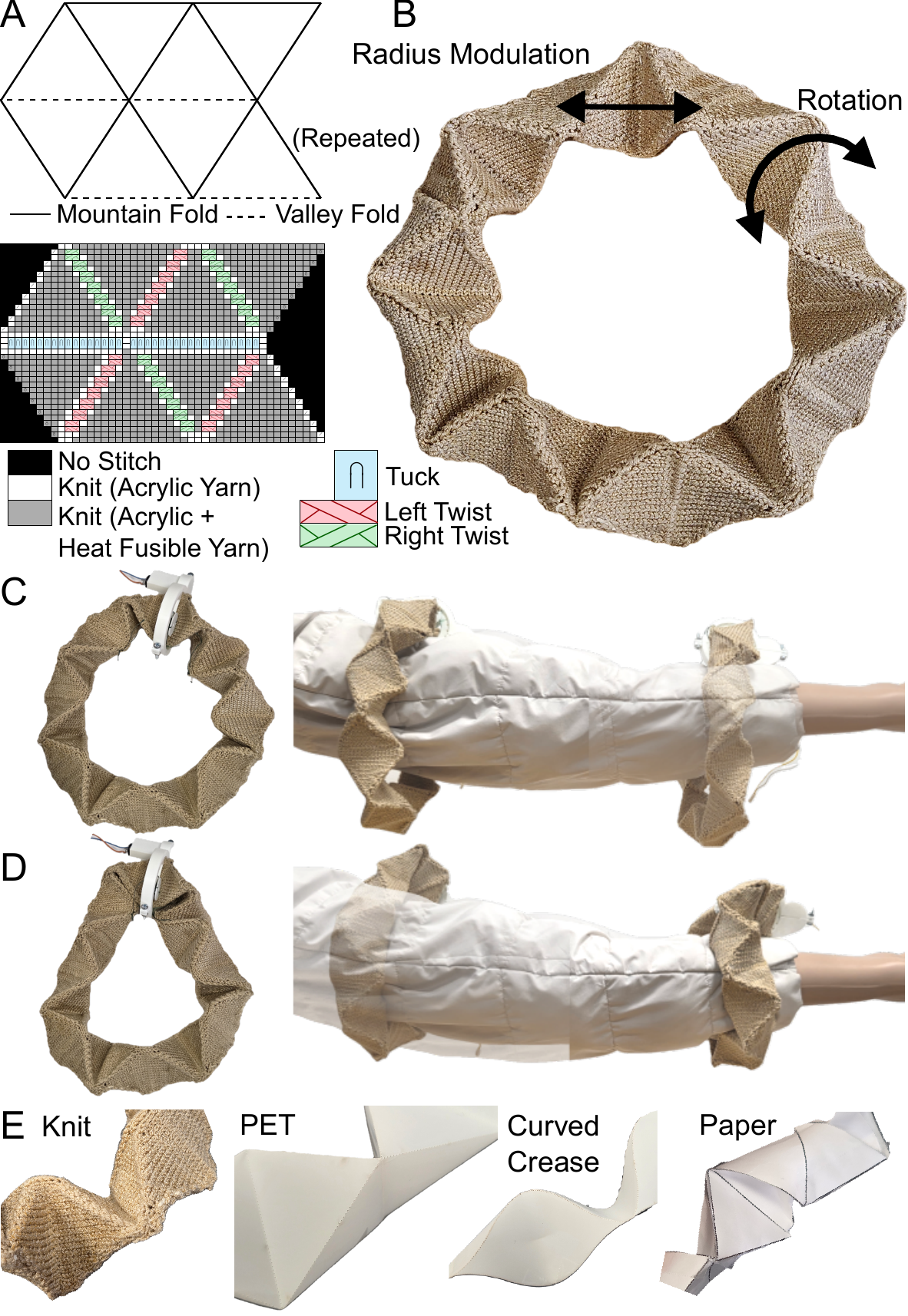}
    \caption{
    Knitted kaleidocycle robot.
(A) Origami pattern (top) and corresponding knit design (bottom) of one kaleidocycle unit. For an $N$-sided kaleidocycle, this unit is repeated $N$ times.
(B) Photograph of the knitted kaleidocycle with its two degrees of freedom: continuous rotation and radius modulation by selectively folding specific sides.
(C) (Left) Kaleidocycle without selective folding. (Right) Fully opened Kaleidocycle worn on the mannequin arm.
(D) (Left) Kaleidocycle with selective folding, resulting in a reduced radius. (Right) Selectively closed Kaleidocycle worn on the mannequin arm.
    (E) Partial view of kaleidocycles made with (i) knitted fabric, (ii) PET thin sheet, (iii) PET thin sheet with curved crease lines, and (iv) paper.
    }
    \label{fig:Fig7}
\end{figure}

A kaleidocycle is a closed chain of hinged tetrahedra that rotates about its central axis \cite{zhang2019origami, tang2018bifurcated, evans2015multistable}. Figure~\ref{fig:Fig7} shows a wearable soft robot that adopts this geometry and is fabricated using the proposed knit-origami method.

As an origami, the kaleidocycle enables continuous rotation while preserving an open axial channel. This behavior is not achievable with a simple torus because the inner and outer radii differ and the surfaces would need to invert during rotation. With more than six tetrahedral units, additional degrees of freedom allow rotation even when some faces are closed. Using this property, an eight-sided knitted kaleidocycle travels along slender bodies while adjusting its radius within a set range.

As a knitted structure, the device remains soft and comfortable while providing sufficient distributed stiffness for reliable folding and rotation. Figure~\ref{fig:Fig7}(E) compares knit, plastic sheet, and paper versions of Kaleidocycle. Plastic sheets were cut along outlines and perforated along crease lines, creating sharp edges and high stiffness that reduce comfort. Using curved creases mitigates sharp vertices, but edges remain sharp. Paper is overly compliant and tends to twist under actuation, even when reinforced with clear tape. In contrast, the knitted version keeps compliance and comfort while selective stiffening prevents undesired deformation.

Actuation consists of two parts. To produce rotation, one edge is driven by a 5 V DC motor through a 10:30 gear train. Radius modulation uses a tendon-driven method in which two tendons routed to three vertices are reeled on a spool to change their effective length. This configuration supports an on-arm wearable device that locomotes along the arm while maintaining an internal channel.

Overall, combining knit and origami yields a geometry with continuous rotation and controllable radius, together with materials that support comfort, durability, and robust operation. This approach is promising for wearable devices that require complex structure reconfiguration while maintaining wearability.

\section{Conclusions}
This work demonstrates a design and fabrication framework that introduces origami to programmed knit to realize soft robotic systems that are easy to manufacture, reconfigurable, and wearable. We present three key contributions.

First, we establish a design method for encoding origami fold patterns into knitted textiles through strategic placement of stitches and heat fusible yarn. This addresses critical manufacturing challenges in origami-based soft robotics, such as delamination, interfacial failure, and balancing compliance with stiffness. The inherently entangled structure of knit enables seamless multimaterial integration within a single continuous fabric, eliminating the need for interfacial bonding. Furthermore, spatially programmed stitch patterns allow direct tuning of local stiffness and anisotropy, enabling the precise design of directional fold.

Second, we experimentally validate the framework through systematic characterization of folding behavior. We measured folding moments at prescribed fold angles and compared samples with/without stitch patterning and with/without heat fusible yarn. Results confirm that the proposed stitch improved the directionality ratio $R$ by reducing forward folding moments while increasing backward folding resistance. This enables prescribed folding without unintended deformations like snap-through. Additionally, heat fusible yarn enhances geometric fidelity by reducing edge curling and prevents out-of-plane deformations by stiffening the panels.

Third, we demonstrate the framework through successful reproduction of complex origami tessellations, including Miura-ori, Kresling, and Yoshimura patterns. We further present a wearable soft robot, a knitted Kaleidocycle. The wearable robot achieves continuous self-propelled rotation around slender bodies while maintaining structural integrity under load and conforming comfortably to the body. The programmed knit design prevents undesired twisting and buckling without sacrificing the compliance necessary for safe human interaction.

Ultimately, our framework demonstrates that by integrating geometric reconfigurability with material programmability and scalable manufacturing, knitted origami holds significant potential for novel design in wearable robotics, assistive devices, and human-robot interaction systems.

\ifblindreview

\else
  \section*{ACKNOWLEDGMENT}
This work was supported in part by the National Science Foundation grant \#2301355. We acknowledge Dr. Cosima du Pasquier for insightful comments and helpful discussions. We acknowledge Sheza S. Saiyed and Fernando Gonzalez as undergraduate researchers for contributions during the early stages of this project.
\fi

\bibliographystyle{IEEEtran}
\bibliography{Reference}

\end{document}